\documentclass{article}

\usepackage[final,nonatbib]{neurips_2020}

\usepackage[utf8]{inputenc} 
\usepackage[T1]{fontenc}    
\usepackage{hyperref}       
\usepackage{url}            
\usepackage{booktabs}       
\usepackage{amsfonts}       
\usepackage{nicefrac}       
\usepackage{microtype}      

\usepackage{graphicx}

\usepackage{hyperref}

\usepackage{amssymb}
\usepackage{amsmath,bm,mathrsfs,amsfonts,algorithm}

\usepackage[noend]{algpseudocode}

\usepackage{listings}
\makeatletter
\gdef\lst@numberfirstlinefalse{\global\let\lst@ifnumberfirstline\iffalse}
\lst@AddToHook{Init}{\global\let\lst@ifnumberfirstline\iftrue}
\makeatother

\usepackage{soul}

\usepackage{tikz}
\usetikzlibrary{positioning,fit}
\usetikzlibrary{decorations.text}
\usetikzlibrary{arrows}
\usetikzlibrary{shapes}

\tikzstyle{atom}  =  [circle, scale=1.7, circle, shading=ball]

\tikzset{my label/.style args={#1:#2}{
  append after command={
    (\tikzlastnode.center) node [#1] {#2}
    }
  }
}

\tikzstyle{edgenode}  =  [thin, draw=black, align=center,fill=white,font=\small]
\tikzstyle{edgeweight}  =  [near start, above, sloped,outer sep=3pt, inner sep=1pt ,fill=white,font=\large]
\tikzstyle{gredge}  =  [outer sep=6pt, inner sep=1pt ,fill=white]
\tikzstyle{gredge1}  =  [outer sep=2pt, inner sep=1pt ,fill=white]

\tikzstyle{kappa} = [
  draw, scale = 1.5,  minimum size=1cm, circle, rounded corners,shading=radial,inner color=white,
  minimum height=1cm,
  align=center
  ]
  
\tikzstyle{lambda} = [
  draw, scale = 1.5,  minimum size=1cm, circle, rounded corners,shading=radial,inner color=white,
  minimum height=1cm,
  align=center
  ]

\newtheorem{example}{Example}

\usepackage{xcolor}

\usepackage{color}

\definecolor{mygreen}{rgb}{0,0.6,0}
\definecolor{mygray}{rgb}{0.5,0.5,0.5}
\definecolor{kwrds}{rgb}{0.99,0.0,0.0}
\definecolor{backcolour}{rgb}{0.95,0.95,0.92}

\newcommand{\weight}[1]{\textcolor{mygreen}{\mathbf{#1}}}
\newcommand{\scalar}[1]{\textcolor{mygreen}{#1}}

\newcommand{\conj}{\textcolor{kwrds}{\scalebox{1.2}{, }}}

\newcommand{\impl}{\textcolor{blue}{ {\scalebox{1.5}{{:-}}} }}

\newenvironment{centeredornot}
    {\begin{center}
    \begin{tabular}{c}
    }
    {
    \end{tabular} 
    \end{center}
    }
    
\tikzstyle{atom}  =  [circle, scale=1.7, circle, shading=ball]

\tikzset{my label/.style args={#1:#2}{
  append after command={
    (\tikzlastnode.center) node [#1] {#2}
    }
  }
}

\tikzstyle{edgenode}  =  [thin, draw=black, align=center,fill=white,font=\small]
\tikzstyle{edgeweight}  =  [near start, above, sloped,outer sep=3pt, inner sep=1pt ,fill=white,font=\large]
\tikzstyle{gredge}  =  [outer sep=6pt, inner sep=1pt ,fill=white]
\tikzstyle{gredge1}  =  [outer sep=2pt, inner sep=1pt ,fill=white]

\tikzstyle{kappa} = [
  draw, scale = 1.5,  minimum size=1cm, circle, rounded corners,shading=radial,inner color=white,
  minimum height=1cm,
  align=center
  ]
  
\tikzstyle{lambda} = [
  draw, scale = 1.5,  minimum size=1cm, circle, rounded corners,shading=radial,inner color=white,
  minimum height=1cm,
  align=center
  ]

\title{Learning with Molecules \\beyond Graph Neural Networks}

\author{%
Gustav Sourek\\
    Czech Technical University\\
  \texttt{souregus@fel.cvut.cz} \\
  \And
        Filip Zelezny       \\ 
        Czech Technical University\\
        \texttt{zelezny@fel.cvut.cz}
  \And
        Ondrej Kuzelka \\
        Czech Technical University\\
        \texttt{kuzelon2@fel.cvut.cz}
}

\begin{document}

\maketitle


\section{Introduction}

Deep learning has registered a tremendous success in the recent past. However, most of the applications are still limited to data in the form of fixed-size feature vectors (tensors).
There are nevertheless many domains where the learning examples are highly structured and do not succumb themselves easily to the precanned form of numeric tensors. 
Particularly, important problems arise around molecular data, which can be understood as attributed graphs of atoms connected via chemical bonds.


To address the problem of deep learning from such structured data (i.e. without preprocessing into feature vectors) Graph Neural Networks (GNNs) have been proposed~\cite{scarselli2008graph,wu2020comprehensive}. GNNs can be viewed as a continuous, differentiable version of the famous Weisfeiler-Lehman (WL) label propagation algorithm used for graph isomorphism refutation checking~\cite{weisfeiler2006construction}. In GNNs however, instead of discrete labels, a continuous node representation (embedding) is being successively propagated into nodes' neighborhoods, and vice versa for the corresponding gradient updates, which can be derived w.r.t. some learning target, such as atom or molecule classification. This paradigm has recently become highly popular~\cite{zhou2018graph}. Nevertheless, there are still considerable limitations to this class of models, stemming from the limited expressiveness of the WL test which is only based on the immediate neighborhood information gathered in each iteration~\cite{xu2018powerful,morris2019weisfeiler}. Consequently, information about more complex relational substructures, such as atom rings in molecules, cannot be properly extracted.

In this paper we demonstrate a deep learning framework which is inherently based in the highly expressive language of relational logic, enabling to, among other things, capture arbitrarily complex graph structures. We show how GNNs and similar models can be easily covered in the framework by specifying the underlying propagation rules in the relational logic. The declarative nature of the used language then allows to easily modify and extend the propagation schemes into complex structures, such as the molecular rings which we choose for a short demonstration in this paper.

\section{Lifted Relational Neural Networks}


We follow up on the framework of Lifted Relational Neural Networks (LRNNs)~\cite{sourek2018lifted} allowing for \textit{templated} modeling of diverse neural architectures oriented towards relational data\footnote{the framework is available at \url{https://github.com/GustikS/NeuraLogic}}. It can be understood as a differentiable version of simple Datalog~\cite{unman1989datalog} programming, where the learning templates, encoding various neuro-relational architectures, take the form of \textit{parameterized} logic programs~\cite{bratko2001prolog}.
It differs from the commonly used frameworks (e.g. TensorFlow) in its declarative nature, which is particularly useful for relational learning problems, such as learning with molecules~\cite{sourek2020beyond}.


\subsection{Learning Examples}

In LRNNs, the learning examples are commonly represented with weighted ground logical facts. A learning example is then a set ${E} = \{(V_1, e_1),\dots,(V_j,e_j)\}$, where each $V_i$ {is} a real-valued tensor and each $e_i$ is a ground fact, i.e. expression of the form
\begin{centeredornot}
\begin{lstlisting}[mathescape=true]
$\weight{V_1}$:: p$_1(c^1_1,\dots,c^1_q)$.
$\dots$
$\weight{V_j}$:: p$_n(c^n_1,\dots,c^n_r)$.
\end{lstlisting}
\end{centeredornot}
where $p_1,\dots,p_n$ are predicates with corresponding arities $q,\dots,r$, and $c_i^j$ are arbitrary constants. Note that this representation allows to encode arbitrary information about atoms (e.g. $\scalar{2.35}$\textcolor{red}{::}$ionEnergy(c_1,level_2$)) and their conformations (e.g. $\scalar{[2.7, -1]}$\textcolor{red}{::}$bond(c_1,o_2)$) in molecules, as demonstrated in the left part of Fig.~\ref{fig:template}.

\subsection{Learning Template}

The learning program, i.e. the declarative \textit{template}, is then set of parameterized rules $\mathcal{T} = \{\alpha_i, \{W^{\alpha_i}_j\}\} = \{(W^i, c) \leftarrow (W_1^i, b_1), \dots, (W_k^i, b_k)\}$, i.e. expression of the form
\begin{centeredornot}
\begin{lstlisting}[mathescape=true]
$\weight{W^1}$ :: h$_1^1$($\dots$) $\impl$ $\weight{W^1_{1}}$ : b$^1_1$($\dots$) $\conj$ ... $\conj\weight{W^1_j}$ : b$^1_i$($\dots$).
$\weight{W^2}$ :: h$_1^2$($\dots$) $\impl$ $\weight{W^2_{1}}$ : b$^2_1$($\dots$) $\conj$ ... $\conj\weight{W^2_k}$ : b$^2_j$($\dots$).
$\dots$
$\weight{W^n}$ :: h$_p^q$($\dots$) $\impl$ $\weight{W^n_{1}}$ : b$^n_1$($\dots$) $\conj$ ... $\conj\weight{W^n_l}$ : b$^n_k$($\dots$).
\end{lstlisting}
\end{centeredornot}
where h$_i^j$'s and b$_i^j$'s are predicates forming positive, not necessarily different, literals, and $\weight{W_i^j}$'s are the associated tensors (also possibly reused in different places). 
Note that the template does not have to encode a particular model or knowledge about the problem. Instead, it can merely encode a generic mode of computation such as, for instance, the GNN propagation scheme.





\begin{example} \textbf{(GNN)}
\label{ex:template}
Consider a simple template for learning with molecules, encoding a generic idea that the representation ($h(.)$) of a chemical atom (e.g. $a(o_1)$) is dependent on the atoms adjacent to it. Given that a molecule can be represented by the set of contained atoms (e.g. $a(h_1)$) and bonds between them (e.g. $b(h_1,o_1)$), we can encode this idea by a following rule (the $gnn$ rule):
\begin{centeredornot}
\begin{lstlisting}[mathescape=true]
$\weight{W_{h_1}}$ ::h(X) $\impl$ $\weight{W_a}$ :a(Y)$\conj\weight{W_b}$ : b(X,Y).
\end{lstlisting}
\end{centeredornot}
where $X,Y$ are free variables. Moreover, one might be interested in using the representation of all atoms ($h(X)$) for deducing the representation of the whole molecule ($q$), for which we can write
\begin{centeredornot}
\begin{lstlisting}[mathescape=true]
$\weight{W_q}$ :: q $\impl$ $\weight{W_{h_2}}$ :h(X).
\end{lstlisting}
\end{centeredornot}
\end{example}

\subsection{Computation Graphs Defined by LRNNs}
\label{sec:semantics}

Here we briefly outline the mapping from a learning template $\mathcal{T}$ and example $E_l$ onto a (differentiable) computation graph $\mathcal{G}_l$.
For that, we take $\mathcal{N}_l = \mathcal{T} \cup E_l$ and {construct} the least Herbrand model $\overline{\mathcal{N}_l}$ of {$\mathcal{N}_{l}$}, which can be done using standard theorem proving techniques~\cite{gallier2015logic}. Next we project the derived logical constructs onto specific node types in the computation graph $\mathcal{G}_l$, the structure of which, broadly speaking, reflects the structure of the derived proof paths. An overview of the node types and their correspondence to common GNN terminology is in Tab~\ref{tab:overviewNNconstruction}. For further details we refer to~\cite{sourek2018lifted}.



\begin{table*}[t]

    \centering
    \begin{tabular}{llllc}
        GNN terminology      & Logical constructs &     Type of node &  Notation  \\
        \hline
        Input data        & Ground fact $h$ &   Fact node  & $F_{(h,\Vec{w})}$   \\
        Convolution  & Ground rule's $\alpha\theta$ body &   Rule node  &   $R_{(W_0^c c\theta \leftarrow W_1^{\alpha} b_1\theta\wedge\dots\wedge W_k^{\alpha} b_k\theta)}^{c\theta}$        \\
        Pooling  & Rule's $\alpha$ ground head $h$ &    Aggregation node &   $\textit{G}_{(W_0^c c \leftarrow W_1^{\alpha} b_1 \wedge \dots \wedge W_k^{\alpha} b_k)}^{h=c\theta_i}$        \\
        Combination       & Ground atom $h$ &   Atom node  &   $A_h$ \\
    \end{tabular}
    \caption{Correspondence between the common GNN terminology and LRNN transformation of a logical ground (Herbrand) model into a computation graph.}
        \label{tab:overviewNNconstruction}
\end{table*}



\begin{figure*}[t!]
\centering
\resizebox{1.0\textwidth}{!}{
	\input{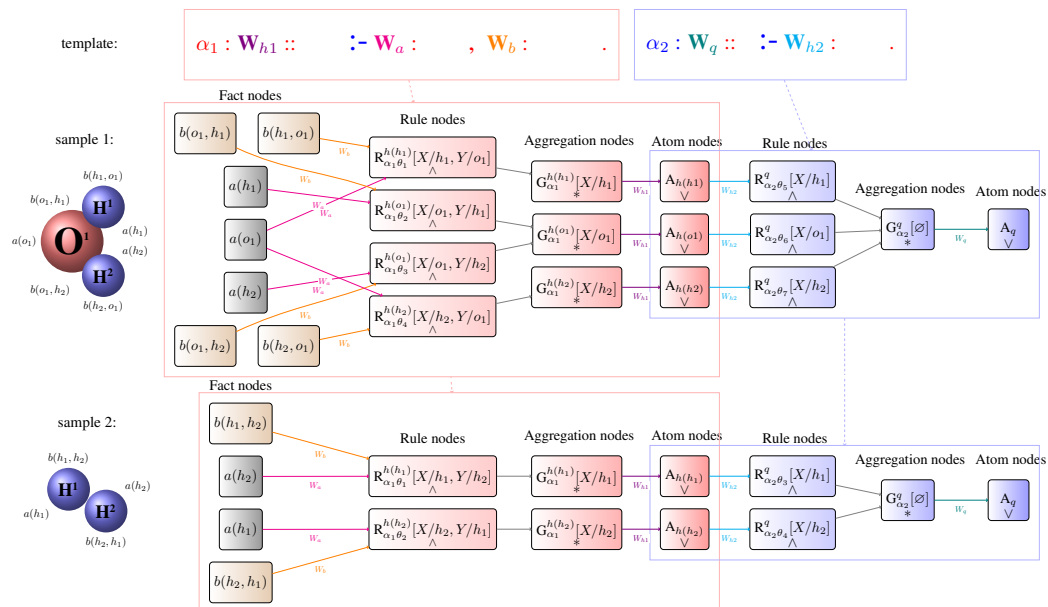}
}
\caption{A simple LRNN template with 2 rules described in Ex. 1. Upon receiving 2 molecules, 2 neural computation graphs get created, as prescribed by the LRNN semantics (Sec.~\ref{sec:semantics}).}
\label{fig:template}
\end{figure*}

\begin{example} \textbf{(GNN cont'd)}
\label{ex:ground}
Let us extend the template from Ex.~\ref{ex:template} with descriptions of two example molecules of hydrogen and water. The derived computation graphs are then displayed in Fig.~\ref{fig:template}.
\end{example}

\section{Extending GNNs with Rings}

We have introduced how standard GNNs can be easily encoded in LRNNs in Ex.~\ref{ex:template}. Note that the example templates discussed in this paper are actual code that can be run very efficiently. For a more detailed description and comparison with existing GNN frameworks we refer to~\cite{sourek2020beyond}. Now we provide a short demonstration on how to reach beyond GNNs with the molecular rings. Using the language of relational logic, a ring can be easily defined as a crisp pattern based on the existing bonds as

\begin{centeredornot}
\begin{lstlisting}
_ring$_6(A,\dots,F)~~\impl$ $$bond$(A,B)\conj\dots\conj$$$bond$(E,F)\conj$$$bond$(F,A)$.
\end{lstlisting}
\end{centeredornot}
\vspace{-0.3cm}
A distributed representation of a ring can then be declared by aggregating all the contained atoms:
\begin{centeredornot}
\begin{lstlisting}
$\weight{W_r}$::ring$^{(n)}_6(A,\dots,F)~~\impl$ _ring$_6(A,\dots,F)\conj\weight{W_a}$:atom$^{(n)}(A)$$\conj\dots\conj\weight{W_f}$:atom$^{(n)}(F)$.
\end{lstlisting}
\end{centeredornot}
\vspace{-0.3cm}
These can be further stacked hierarchically, replacing the crisp $\_\text{ring}_6$ pattern by the respective distributed representation $\text{ring}^{(n-1)}_6$ from the preceding layer. Similarly, we can then also propagate the representation from the ring back into all the contained atoms by simply adding the following rule
\begin{centeredornot}
\begin{lstlisting}
$\weight{W_{a'}}$::atom$^{(n)}(A)$ $\impl$ $\weight{W_{r'}}$: ring$^{(n-1)}_6(A,\dots,F$).
\end{lstlisting}
\end{centeredornot}
Note that each ring pattern will be automatically instantiated in all possible revolutions, and so it is enough to specify propagation into the ``first'' atom ($A$) only. We further refer to this template as $rings$. We note that it is a good practice to add some basic transformation to produce at least some output should a molecule contain no rings at all, e.g. simply aggregating all pairs of connected atoms:
\begin{centeredornot}
\begin{lstlisting}
$\weight{W_q}$::molecule $\impl$ $\weight{W_1}$:atom($A$)$\conj$$\weight{W_2}$:atom($B$)$\conj$bond$(A,B$).
\end{lstlisting}
\end{centeredornot}
\vspace{-0.3cm}
Naturally, this can be further combined with the standard GNN propagation scheme (Ex.~\ref{ex:template}), where we aggregate the direct neighbor representations:
\begin{centeredornot}
\begin{lstlisting}
$\weight{W_h}$::atom$^{(n)}(X)$ $\impl$ $\weight{W_{1'}}$:atom$^{(n-1)}(Y)\conj$bond($X,Y$).
\end{lstlisting}
\end{centeredornot}
\vspace{-0.3cm}
In each layer $n$, the representation of an atom will thus get updated based on the neighbors as well as all the atoms occupying the same rings. The particular contributions to the representation update are then parameterized separately. We further refer to this joint template as ${rings+gnn}$.

\begin{figure}
    \centering
    \resizebox{1.0\textwidth}{!}{
    \includegraphics{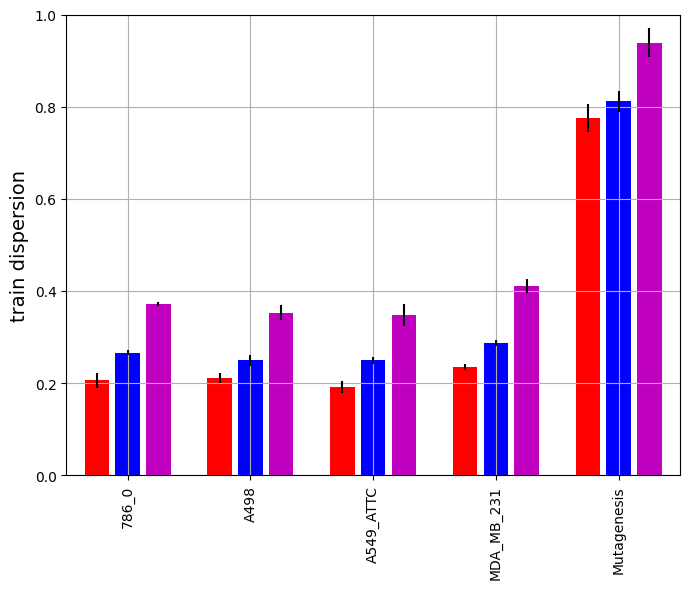}
    \includegraphics{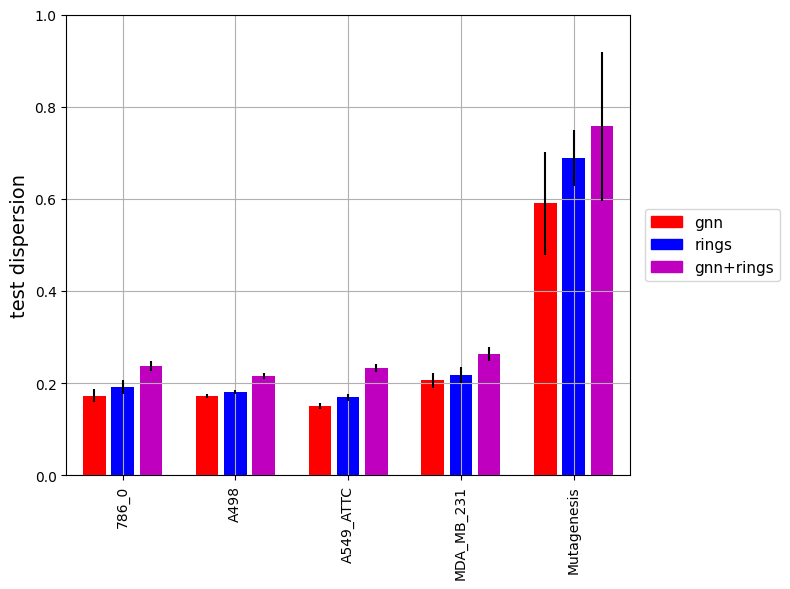}
    }
    \caption{Comparison of the ${gnn}$, ${rings}$, and the joint ${gnn+rings}$ learners across several molecule classification datasets w.r.t. training (left) and testing (right) dispersion (higher is better).}
    \label{fig:rings}
\end{figure}

\subsection{Experiments}

For the experiments we encoded both 5 and 6 rings, added 3 layers for both the $\sf{gnn}$ and $\sf{rings}$ embedding propagation, and set all the learnable tensor to randomly initialized $3\times3$ matrices. We then evaluated the architectures across several datasets for molecule classification~\cite{ncigi,mutagenesis,PTC}, ranging from 188 to app. 3500 molecules. We note we only used the basic information on atom types and their conformations. We left all the hyperparameters of the framework on their defaults, learning with $tanh$ activations, $avg$ aggregations, and $ADAM$ optimizer, trained for $2000$ steps, and evaluated with a 5-fold cross-validation. The results are displayed in Fig.~\ref{fig:rings}. We can see that the ${rings}$ template indeed performs better than the basic $gnn$ learner, while both benefit from their mutual combination.


\section{Discussion}

We have demonstrated a deep relational learning framework of LRNNs~\cite{sourek2018lifted} on a small learning scenario designed to extend the basic GNN idea to capture molecular rings. Note that all we had to do was to declare a couple of simple rules specifying (i) what a ring is and (ii) how to update representation of its contained atoms. The derivation of the particular dynamic computation graphs, encoding the respective propagation scheme for each of the differently structured molecules, the corresponding gradients, training and evaluation were then all performed completely automatically. This allows for quick prototyping of diverse relational modelling ideas, where one can stack complex structural patterns to be trained end-to-end with gradient descend. In the particular showcase just demonstrated, we were then able to quickly evaluate the contribution of atom rings to basic GNNs.



\bibliography{references}
\bibliographystyle{plain}

\end{document}